\documentclass[lettersize,journal]{IEEEtran}
\usepackage{amsmath,amsfonts}
\usepackage{algorithmic}
\usepackage{array}
\usepackage[caption=false,font=normalsize,labelfont=sf,textfont=sf]{subfig}
\usepackage{textcomp}
\usepackage{stfloats}
\usepackage{url}
\usepackage{verbatim}
\usepackage{graphicx}
\usepackage[numbers]{natbib}
\usepackage{orcidlink}
\usepackage[flushleft]{threeparttable}
\usepackage{multirow}
\usepackage{booktabs}
\usepackage{tabularx}
\usepackage{amsmath}

\newcolumntype{C}{>{\Centering}X} % define centered version of 'X' col. type
\def\BibTeX{{\rm B\kern-.05em{\sc i\kern-.025em b}\kern-.08em
    T\kern-.1667em\lower.7ex\hbox{E}\kern-.125emX}}
\usepackage{balance}
\begin{document}
\title{Street-Level Geolocalization Using Multimodal Large Language Models and Retrieval-Augmented Generation}
\author{Yunus Serhat Bıçakçı\,\orcidlink{0000-0002-7288-9959}, Joseph Shingleton\,\orcidlink{0000-0002-1628-3231}, Anahid Basiri\,\orcidlink{0000-0002-2399-1797}
\thanks{\it(Corresponding author: Yunus Serhat Bıçakçı.)}
\thanks{Yunus Serhat Bıçakçı is with the Vocational School of Social Sciences, Marmara University, İstanbul 34865, Türkiye and also with the Geospatial Data Science Group, School of Geographical \& Earth Sciences, University of Glasgow, Glasgow G12 8QQ, Scotland, UK, (e-mail: yunus.serhat@marmara.edu.tr).}
\thanks{Joseph Shingleton and Anahid Basiri are with the Geospatial Data Science Group, School of Geographical \& Earth Sciences, University of Glasgow, Glasgow G12 8QQ, Scotland, UK.}}

% \markboth{IEEE Journal of Selected Topics in Applied Earth Observations and Remote Sensing,~Vol.~17,~2024}{}

\IEEEpubid{© 2025 The Authors. This work is licensed under a Creative Commons Attribution 4.0 License. For more information, see https://creativecommons.org/licenses/by/4.0/}

\maketitle
\begin{abstract}
Street-level geolocalization from images is crucial for a wide range of essential applications and services, such as navigation, location-based recommendations, and urban planning. With the growing popularity of social media data and cameras embedded in smartphones, applying traditional computer vision techniques to localize images has become increasingly challenging, yet highly valuable. This paper introduces a novel approach that integrates open-weight and publicly accessible multimodal large language models with retrieval-augmented generation. The method constructs a vector database using the SigLIP encoder on two large-scale datasets (EMP-16 and OSV-5M). Query images are augmented with prompts containing both similar and dissimilar geolocation information retrieved from this database before being processed by the multimodal large language models. Our approach has demonstrated state-of-the-art performance, achieving higher accuracy compared against three widely used benchmark datasets (IM2GPS, IM2GPS3k, and YFCC4k). Importantly, our solution eliminates the need for expensive fine-tuning or retraining and scales seamlessly to incorporate new data sources. The effectiveness of retrieval-augmented generation-based multimodal large language models in geolocation estimation demonstrated by this paper suggests an alternative path to the traditional methods which rely on the training models from scratch, opening new possibilities for more accessible and scalable solutions in GeoAI.
\end{abstract}


\section{Introduction}
\IEEEPARstart{S}{treet-level} geolocalization refers to the task of determining the precise geographic location of a scene from one single image \cite{vo2017revisiting}. This capability underpins numerous applications and services such as humanitarian aid and disaster relief \cite{hernandez2019using, Suwaileh2022}, mis/dis-information detection \cite{zhao2017true, kordopatis2017geotagging}, automated mapping \cite{louro2022geolocalization, gupta2016ubiquitous}. The growing popularity of social media platforms and the widespread availability of cameras in smartphones have generated vast amounts of user-created images, making the need for accurate and scalable geolocation methods more urgent than ever. \cite{ACHARYA2022104152, cui2023analysing, zhang2024multi}. However, the street view imagery (SVI) can pose challenges to currently existing methods due to their viewpoint variance, cluttered urban environments, and sometimes limited metadata \cite{biljecki2021street, 2024_global_streetscapes, astruc2024openstreetview5mroadsglobalvisual}. Despite recent progress, current methods still grapple with issues such as sensitivity to environmental factors \cite{rodrigues2021these, lisovski2012geolocation, kinnari2022season}, sparsely labeled data \cite{biljecki2021street} that generalize effectively across diverse regions. Additionally, high computational costs and domain-specific challenges often prevent conventional image-based geolocation methods from achieving robust street-level accuracy in real-world scenarios.

Recent developments in large language models (LLMs) \cite{brown2020languagemodelsfewshotlearners}, multimodal large language models (MLLMs) \cite{Yin_2024} and Retrieval-Augmented Generation (RAG) \cite{lewis2021retrievalaugmentedgenerationknowledgeintensivenlp, gao2024retrievalaugmentedgenerationlargelanguage} offer promising solutions to these challenges \cite{10433480, xu2024genaipoweredmultiagentparadigmsmart} in geospatial artificial intelligence (GeoAI). The process—initially driven by computer vision models \cite{hays2008im2gps, weyand2016planet}, advanced with the advent of language–image pre-training \cite{radford2021learning, zhai2023sigmoid, li2023blip, oquab2023dinov2}, and most recently, predictions have been continually enhanced through the use of MLLMs. These models, already proficient across various tasks, are now capable of achieving high-accuracy geolocalization. In particular, RAG methods \cite{gao2024retrievalaugmentedgenerationlargelanguage} can offer a powerful means of task-specific improvement by iteratively providing the model with existing data, thereby reducing both costs and time expenditure associated with large-scale re-training.

In this study, we explore how the integration of MLLMs and RAG techniques can further enhance geolocation estimation in SVI, and we present the state-of-the-art results and methodologies we have achieved.

\IEEEpubidadjcol

Following the advent of the transformer architecture \cite{DBLP:journals/corr/VaswaniSPUJGKP17}, we have witnessed rapid advancements in language modeling \cite{devlin2019bertpretrainingdeepbidirectional, liu2019robertarobustlyoptimizedbert, brown2020languagemodelsfewshotlearners, chowdhery2022palmscalinglanguagemodeling, touvron2023llama}. Subsequently, with the integration of image and text pairings into a unified representation space \cite{radford2021learning}, MLLMs have emerged as a powerful tool for vision-language tasks \cite{openai2024gpt4technicalreport, liu2023visualinstructiontuning, chen2024internvlscalingvisionfoundation, yao2024minicpmvgpt4vlevelmllm, bai2023qwenvlversatilevisionlanguagemodel, dubey2024llama3herdmodels}. LLMs and MLLMs have become central to advancements in artificial intelligence, and their capabilities can be harnessed in GeoAI. The success of these models lies in their ability to understand and generalize both text and images. Equipped with the capacity to parse and generate both textual and visual content, MLLMs have achieved significant success in general domains such as image captioning \cite{vinyals2015show, karpathy2015deep}, visual question answering \cite{antol2015vqa}, and cross-modal information retrieval \cite{faghri2017vse++, lu2019vilbert}. However, their success in relatively specific or complex domains like geolocation estimation remains under question \cite{zhou2024img2loc}. This could be due to the inherent nature of MLLMs, which are trained to prioritize generalization over handling highly specific tasks such as geolocation estimation.

To accurately achieve geolocation estimation, models require a wide range of diverse information, such as landmarks \cite{hays2015large} and architectural structures \cite{doersch2015makes}. Nevertheless, purely classification-based and retrieval-based methods have only made incremental improvements in street-level geolocalization accuracy \cite{weyand2016planet, muller2018geolocation, seo2018cplanet, pramanick2022world, clark2023we, zhu2022transgeo, lin2022joint, zhang2023cross}. This may suggest that the succesful performance bottlenecks remain in real-world scenarios. 

While similar methods \cite{zhou2024img2loc} have demonstrated the feasibility of retrieval-enhanced geolocation, this paper introduces a robust hybrid gallery—combining the Extended MediaEval Placing Tasks 2016 Dataset (EMP-16) \cite{larson2017benchmarking, Theiner_2022_WACV} and the OSV-5M dataset \cite{astruc2024openstreetview5mroadsglobalvisual}—along with an open-weights MLLMs. This can address the challenges related to cost constraints, and scales to diverse data sources to provide accurate street-level geolocalization. To do so, we transform these datasets into an embedding space using the SigLIP image encoder \cite{zhai2023sigmoid}, forming a vector database that stores each image’s embedding alongside its geolocation information. Consequently, any new query image can be converted into an embedding via the same encoder, and the geolocation information of similar and dissimilar images is efficiently retrieved using the Faiss library \cite{douze2024faiss, johnson2019billion}. Then we prompt the MLLM with (i) the query image itself, and (ii) the retrieved geolocations from the most similar and most dissimilar images. Thus, the model should be able to execute the location estimation capability with an acceptable precision at the street-level, leveraging both general information about the image and location information from our database through the RAG approach. This approach enables us to compare the street-level location estimation results with benchmark datasets including IM2GPS \cite{hays2008im2gps}, IM2GPS3k \cite{vo2017revisiting}, and YFCC4k \cite{thomee2016yfcc100m}.

This paper makes contributions to GeoAI field, in particular to the task of estimating geolocation for SVI. Below, we summarize our key contributions:

\begin{itemize}
\item{This study demonstrates that high geolocalization estimation accuracy can be achieved using MLLMs and RAG databases.}
\item{The study provides cost and time savings by not performing pre-training and fine-tuning processes often required for other models.}
\item{Our approach exhibits state-of-the-art performance at street-level on three benchmark datasets (IM2GPS, IM2GPS3k, and YFCC4k), and achieves very good results at all other levels.}
\item{We offer performance comparisons with different datasets and models and present comparative tables illustrating which open-weights models excel in geolocation estimation.}
\end{itemize}

\section{Related Work}
While there are many geolocalization technologies and techniques such as GPS \cite{aguilar2012geopositioning}, Wi-Fi signals \cite{zhuang2015autonomous}, or multispectral satellite imagery \cite{rs13193979}, with the availability and popularity of social media platforms as well as miniaturization of smartphone cameras there is a growing need for accurate and scalable street-level image-based geolocalization. Street-level geolocalization refers to the challenge of inferring the geographic location where a single image was captured \cite{vo2017revisiting, 9635972}. 

The use of street-level images can be inherently complex due to factors like temporal variation \cite{rodrigues2021these}, weather conditions \cite{lisovski2012geolocation}, seasonal changes \cite{kinnari2022season}, and urban infrastructures \cite{rs10050661}. Recent advancements in computer vision techniques, coupled with the widespread availability of SVI, have made it possible to map street-level features at high resolutions, facilitating geolocalization for a wide range of applications \cite{biljecki2021street}. However, their scalability and accuracy can be still improved.

These have opened researchers with a wide range of opportunities and challenges to work on. For example, "IM2GPS" by Hays and Efros \cite{hays2008im2gps} is one of the foundational modern efforts in geolocation estimation that tackled one of the key challenges—limited labeled data—by leveraging a dataset of 100,000 images. Their classification-based approach (using support vector machines) illustrated how large-scale image repositories could be utilized for geolocation. While IM2GPS demonstrated the feasibility of learning geographic cues (e.g., terrain, vegetation, landmarks), the method struggled with scenes lacking salient features, highlighting the importance of handling diverse environmental conditions.

Building on this foundation, Tobias Weyand et al. \cite{weyand2016planet} developed a model namely "PlaNet" using deep learning-based convolutional neural networks (CNNs) for geolocation estimation. They approached the geolocation estimation task as a classification problem by dividing the Earth's surface into a series of geographic cells, which served as target classes. Subsequently, they employed CNNs to predict potential locations on the Earth's surface using a continuous probability distribution. This method achieved significantly more accurate results compared to earlier models. Their advancements were further refined in the "CPlaNet" model \cite{seo2018cplanet}, which predicted finer-grained geographic classes through the combinatorial partitioning of multiple geo-class sets. They developed a method that generated a distinct geo-class set for each classifier, incorporating features such as seasonal and weather conditions. This approach resulted in significantly improved performance compared to the previous model.

In 2018, Eric Müller-Budack and collaborators introduced the "ISNs" framework \cite{muller2018geolocation}, which divided the Earth's surface into a hierarchical structure of nested geographic cells. This approach enabled their model to leverage both global and local geographic cues for more precise predictions.

Similarly, the "TransLocator" model \cite{pramanick2022world} approached the problem by segmenting the Earth's surface into numerous geographic cells and assigning each image to one of these cells, treating geolocation estimation as a classification problem. However, TransLocator distinguished itself by employing a Transformer-based architecture, where two parallel Transformer branches interacted through a fusion strategy. This enabled the model to capture fine-grained details within images, further advancing geolocation estimation capabilities.

Clark et al.\cite{clark2023we} introduced GeoDecoder, a model that learns specialized representations of different geographic levels and scene types using a Swin Transformer-based architecture \cite{liu2021swin}. Integrating visual and scene information through a hierarchical cross-attention mechanism, the model captures features specific to each geographic category. However, training on the MediaEval Placing Task 2016 (MP-16) dataset led to lower performance compared to later models, especially in SVI tasks. Additionally, it has been shown that incorporating textual information and language localization can further enhance geolocalization.

PIGEOTTO by Haas et al. \cite{Haas_2024_CVPR}, is built upon the Contrastive Language-Image Pretraining (CLIP) image encoder. It is trained on over 4 million images sourced from the MP-16 dataset and Wikipedia. PIGEOTTO improves geolocalization estimation by leveraging semantic geocells, claiming superior results across a broader geographic scope compared to earlier models. The model demonstrates accuracy improvements at city and country levels, showcasing strong generalization capabilities. However, its performance in SVI tasks remains less competitive compared to our model and other MLLM-based approaches on benchmark datasets. Additionally, the diversity and scale of datasets used for PIGEOTTO’s training, while advantageous for accuracy, can pose computational cost challenges. Despite its architectural similarity to the PIGEON model developed by the same authors, PIGEOTTO achieves better results, supported by the use of distinct datasets in its training process.

GeoCLIP, proposed by Cepeda et al. \cite{vivanco2024geoclip}, employs a CLIP-inspired architecture to align image features with global positioning system (GPS) coordinates. Unlike methods that divide the world into discrete classes, GeoCLIP the Earth as a continuous function through spatial encoding using random Fourier features, effectively creating a hierarchical representation to capture geographic information. The model utilizes a pre-trained CLIP model as its image encoder and introduces a location encoder to map GPS coordinates into the same embedding space as the image features. Leveraging the pre-trained CLIP backbone enables the model to handle text queries, though not as advanced as those processed by MLLMs. While the use of precomputed image features on large datasets reduces computational burden during inference, it imposes significant computational overhead during preprocessing. Additionally, the model's flexibility for post-development modifications is limited, and relying on a single random Fourier feature value may not be sufficient to ensure optimal performance across both small and large geographic areas.

Img2Loc, proposed by Zhou et al.  \cite{zhou2024img2loc}, adopts an approach to image geolocation similar to ours by leveraging MLLMs and image-based RAG. Instead of relying solely on retrieval or classification methods based on image content, the model uses a pre-trained CLIP model to create embeddings from images and their associated geolocations. These features are then utilized to retrieve relevant geographic information from a database of geotagged images. However, the model's results were achieved using an application programming interface (API) that charges per query, raising concerns about the feasibility and scalability of this approach. Additionally, the dataset employed lacked sufficient SVI, leading to suboptimal performance on benchmark datasets for some geographic levels. The reliance on an external, paid API for consistent performance and the dependency on the quality and scope of the RAG database—particularly its insufficient SVI content—limit the robustness and applicability of the method.

In summary, while many models including PlaNet, CPlaNet, ISNs, TransLocator, GeoDecoder, PIGEOTTO, GeoCLIP, and Img2Loc, represent significant progress in image-based geolocation, from coarse-grid classification to transformer architectures and continuous location encodings, there are still several important challenges to address. Specifically:
\begin{itemize}
    \item Model Development: Methods like PIGEOTTO \cite{Haas_2024_CVPR} lead to high computational overhead during training or preprocessing with image datasets. Additionally, making improvements or fine-tuning the models post-development with different images proves to be quite challenging.
    \item SVI Performance: Previous models exhibit subpar results in heavily urbanized or dense city environments, revealing a need for enhanced fine-grained feature extraction.
    \item Reliance on External APIs: Approaches like Img2Loc \cite{zhou2024img2loc} depend on fee-based APIs, which can limit broad applicability and raise feasibility concerns.
\end{itemize}

These limitations highlight the need for more robust, scalable solutions that can effectively handle the challenges of SVI while minimizing reliance on expensive external resources. In light of these, this paper proposes a solution that combines an open-weight MLLM framework with a diverse, locally hosted RAG database to eliminate API dependencies, reduce training overhead, and enhance street-level results, as detailed in the next section. 

\section{Methodology}
In light of these gaps—particularly the challenges of leveraging street-level data, balancing computational costs, and reducing reliance on external APIs — we propose a RAG approach that integrates open-weights MLLMs and includes both similar and dissimilar images geolocation information. This design is flexible allowing expansion to encompass further data modalities. Providing not only positive location cues but also contrasting negative examples, this approach can enhance the accuracy and robustness of geolocation estimations, especially for more complex street-level scenes.

For a given query image, our method for geolocation estimation begins by producing a high-dimensional numerical representation (or embedding) of the image using our chosen encoder. Motivated by the limitations highlighted in the literature review—where purely classification-based or retrieval-only approaches may overlook crucial contextual cues—we adopt a RAG strategy. Specifically, we leverage both similar and dissimilar images from our RAG database (constructed as shown in \autoref{retrieval}) to form an augmented prompt. 

The rationale for including both similar and dissimilar geolocations is twofold. First, similar geolocations provide positive location cues, such as landmarks, architectural styles, or environmental features, which can guide MLLMs toward a probable geolocation. Second, dissimilar geolocations serve as negative contexts,
helping to clarify which visual or geographic features are absent in the query scene. This approach is closely aligned with the contrastive training methods used by many image encoders within MLLMs.

By offering complementary geolocations from both the most similar and most dissimilar matches, this approach enriches the augmented prompt with relevant contextual information, enabling the MLLM to make more accurate geolocation estimations. A detailed step-by-step explanation of how embeddings and retrieved images are utilized is provided in the subsequent subsections.

\subsection{RAG Database Construction and Image Query}

\begin{figure*}[!t]
\centering
\includegraphics[width=\textwidth]{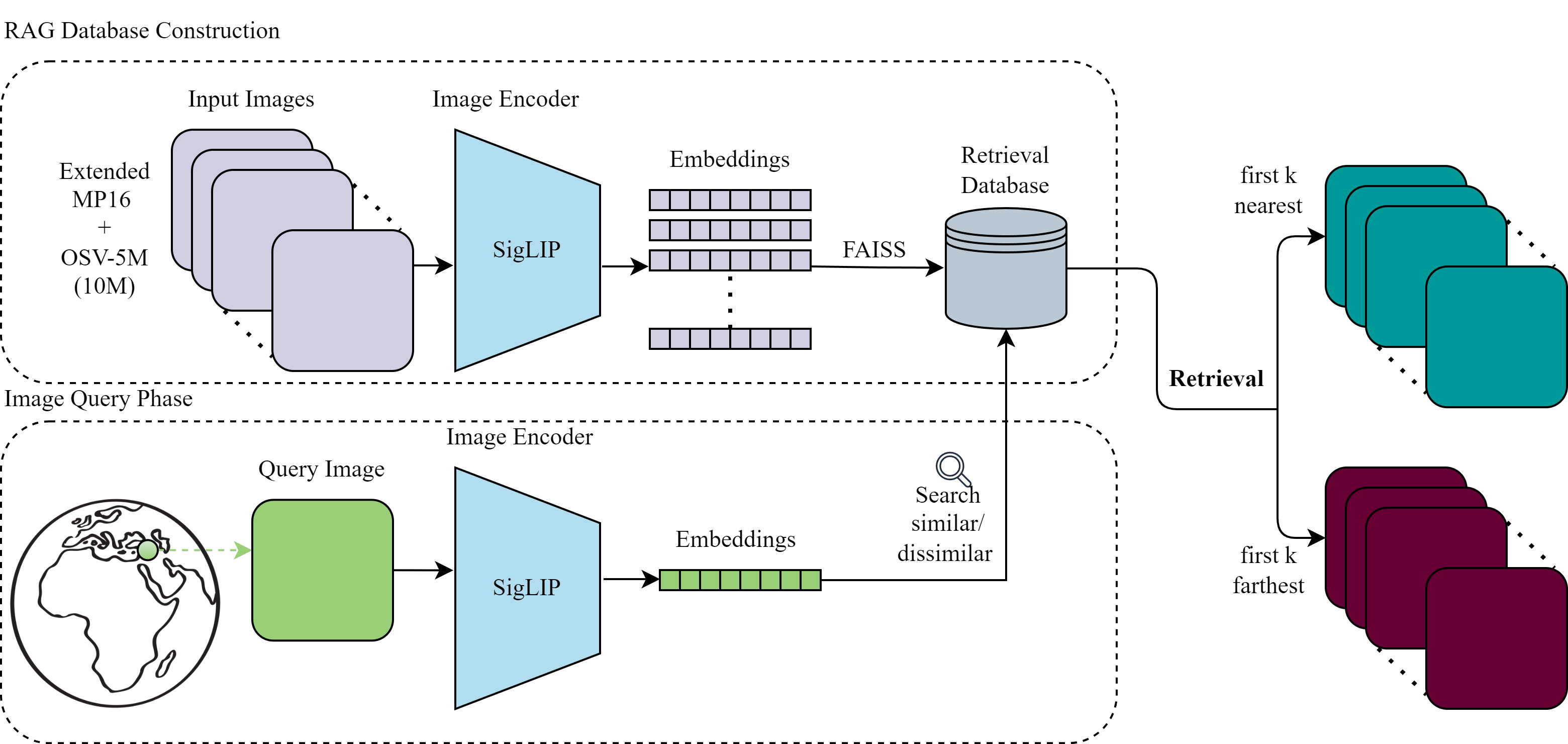}
\caption{The RAG database construction and image query pipeline (adapted from \cite{zhou2024img2loc}).}
\label{retrieval}
\end{figure*}

For the RAG database construction phase, we combine the user-centric EMP-16 dataset \cite{Theiner_2022_WACV}, comprising 4.6 million geo-tagged images captured by ordinary individuals, with the OSV-5M dataset \cite{astruc2024openstreetview5mroadsglobalvisual}, consisting of 5.2 million road street-view geo-tagged images. By merging everyday user-generated photos with structured street-view imagery, we build a hybrid vector database that offers broader coverage of real-world environments and richer location information for evaluating geolocation performance. This approach ensured the inclusion of a greater variety of images from different regions in the vector database, providing the image encoder with a broader range of options to establish close or distant relationships with incoming images. Here, the image encoder is utilized both for creating a vector database from 10 million previously obtained images during the RAG database construction phase as shown in \autoref{retrieval} and image query phase for encoding newly received images to perform searches within the retrieval vector database. 

For our approach, the image encoder plays a pivotal role and serves as a cornerstone of the methodology. The SigLIP model (siglip-so400m-patch14-224), an open-weights release by Google \cite{zhai2023sigmoid}, has been selected as the image encoder. SigLIP was selected as the image encoder not only because it achieved the best results on benchmarks \cite{zhai2023sigmoid} such as ImageNet-1k and COCO R@1, but also because it outperformed alternatives like OpenAI CLIP \cite{radford2021learning}, LAION \cite{ilharco_gabriel_2021_5143773}, and StreetCLIP \cite{haas2023learning} in our own tests conducted. Additionally, once all embeddings are generated through using an image encoder, we utilise FAISS \cite{douze2024faiss}, a tool and library designed to enhance efficiency in vector-based databases, to facilitate searches and manage storage within the database.

After constructing the RAG database, we use the SigLIP image encoder to identify the most similar and most dissimilar neighbors for the incoming image. By "most similar" and "most dissimilar," we refer to finding the nearest and farthest embeddings within the vector space generated by the SigLIP image encoder. In our case, the similarity and dissimilarity are determined based on the L2 distance (Euclidean distance) between vector spaces. This approach allows us to retrieve the location information for as many images as desired, corresponding to the closest and farthest embeddings.

\begin{figure*}[!t]
\centering
\includegraphics[width=\textwidth]{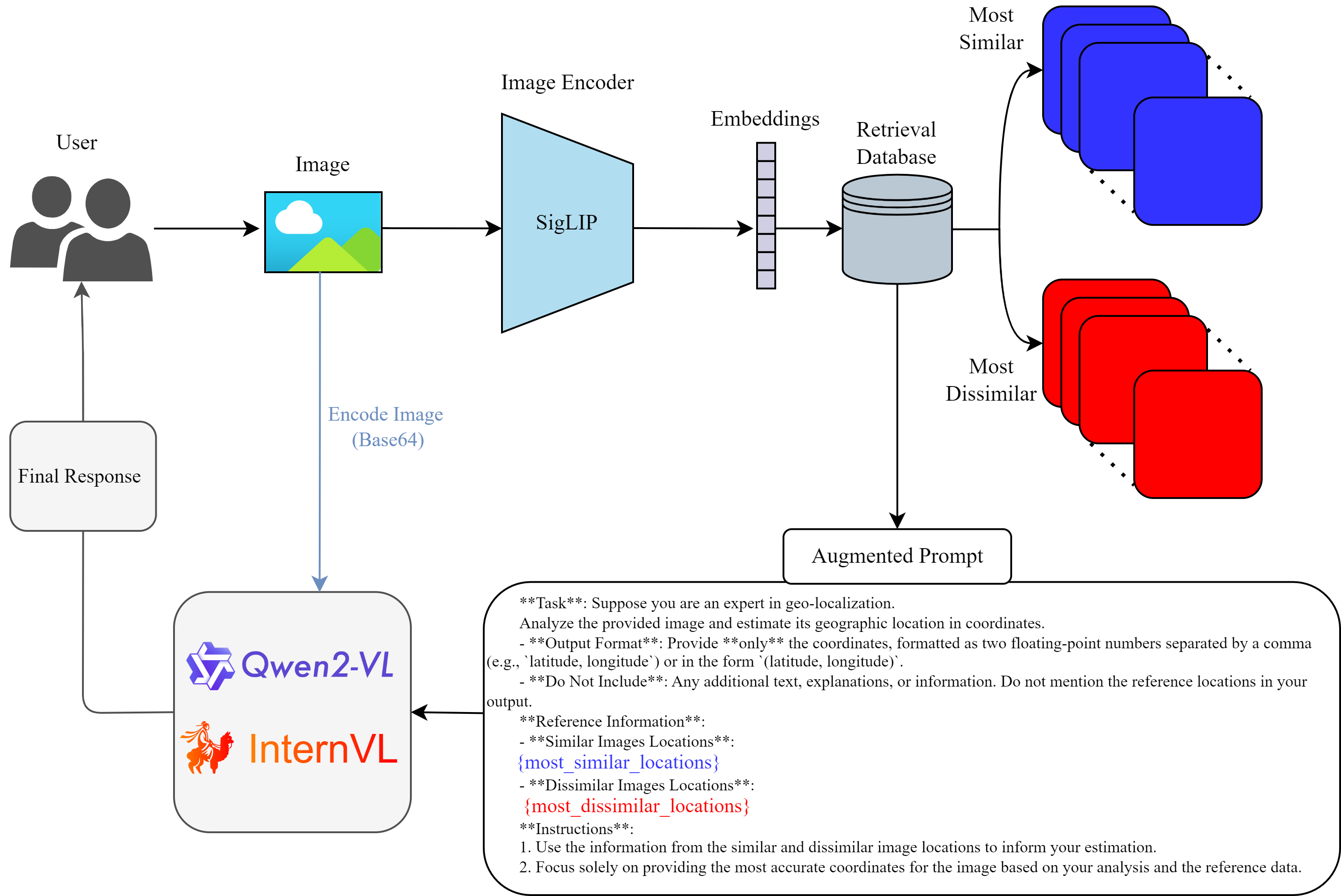} 
\caption{Overall geolocation estimation framework of the proposed method.}
\label{architecture}
\end{figure*}

The location information from the most similar and most dissimilar images is then used to augment the prompt, which serves as input to the MLLMs system. The augmented prompt, comprising the given image and the retrieved geolocations (x, y coordinates), is fed into the MLLMs for geolocation estimation. These steps are illustrated in \autoref{architecture}.

\subsection{Model Selection}

In this study, extensive testing was conducted on numerous open-source MLLMs, including but not limited to the Qwen2-VL \cite{wang2024qwen2vlenhancingvisionlanguagemodels}, InternVL2 \cite{chen2024internvlscalingvisionfoundation, chen2024internvlfar}, Pixtral \cite{agrawal2024pixtral12b}, Llama 3.2 Vision \cite{dubey2024llama3herdmodels}, and Aria \cite{li2024ariaopenmultimodalnative}. The primary reason for selecting these models was their accessibility as open-weight resources on HuggingFace \cite{huggingface}.

The preliminary test results are consistent with previous research \cite{huang2024surveyevaluationmultimodallarge, Yin_2024}. This can promise a higher performance for MLLMs with a higher number of parameters, in geolocation estimation tasks, similar to other tasks \cite{zhang2023m3exam, 10769058, Li_2024_CVPR}. Consequently, certain models mentioned above were excluded from the final analysis to ensure fairness in the comparison. However, these high parameter models have significant graphics processing unit (GPU) resource requirements. To address this issue, the quantized versions of these models, published by their developers using AWQ \cite{lin2024awqactivationawareweightquantization} and GPTQ \cite{frantar2023gptqaccurateposttrainingquantization}, were selected. Ultimately, two of the most successful of our tasks MLLMs —Qwen2-VL-72B-Instruct and InternVL2-Llama3-76B— are utilized.

\subsection{Implementation}

All experiments are conducted utilizing the high-performance computing (HPC) resources of the university affiliated with the authors. Considering the scale of the models employed, we utilized 2 x NVIDIA RTX 6000 Ada GPUs with 48GB GDDR6 memory (18,176 CUDA Cores) on the HPC to facilitate parallel processing and enable efficient model execution. 

The construction of the RAG system was supported by libraries such as PyTorch \cite{paszke2019pytorchimperativestylehighperformance}, Faiss \cite{douze2024faiss}, Pillow \cite{andrew_murray_2024_13935429}, Transformers \cite{wolf2020huggingfacestransformersstateoftheartnatural}, and pandas \cite{reback2020pandas}. Additionally, to run and perform inference with MLLMs, we leveraged the architectures and tools provided by the vLLM \cite{kwon2023efficient} and LMDeploy \cite{2023lmdeploy} libraries.

The selected MLLMs used specific hyperparameter settings to optimize their performance for the geolocation estimation task. For the RAG phase, the prompt generation incorporated both the 16 most similar and the 16 most dissimilar location information. We empirically determined that retrieving the 16 most similar and 16 most dissimilar embeddings yielded the best performance. Fewer neighbors (1, 5, or 10) provided less contrastive information, while larger sets did not yield additional improvements and increased computational overhead. Furthermore, the following hyperparameters were applied for MLLMs: a temperature \cite{openai2024} of "0.1", a top-p value \cite{openai2024} of "0.1", a maximum model length \cite{vllm2024} of "6,000", and a maximum token limit \cite{vllm2024} of "512".

To evaluate the obtained outputs for each image included in the benchmark datasets, the geodetic distance between the location estimated by the MLLMs and the coordinates is calculated using the GeoPy library \cite{geopy}. The computed distance error is then categorised according to thresholds established in previous studies \cite{weyand2016planet, hays2015large, vo2017revisiting}: 1 km for street-level accuracy, 25 km for city-level accuracy, 200 km for region-level accuracy, 750 km for country-level accuracy, and 2,500 km for continent-level accuracy. We used these thresholds to ensure a fair comparison with previous studies conducted on this task and to demonstrate how well our proposed method performs compared to other models.

\section{Results}

\begin{table*}[t]
\centering
\caption{Geolocation estimation accuracy of the proposed method compared to previous methods, across benchmark datasets.}
\label{tab:performance_comparison}
\begin{tabular}{c|l|lllll}
\hline
\multirow{3}{*}{\textbf{Benchmark}}  & \multicolumn{1}{c|}{\multirow{3}{*}{\textbf{Method}}} & \multicolumn{5}{c}{\textbf{Distance (\% @ km)}}      \\
   & \multicolumn{1}{c|}{} & \multicolumn{1}{c}{\textit{Street}} & \multicolumn{1}{c}{\textit{City}}  & \multicolumn{1}{c}{\textit{Region}} & \multicolumn{1}{c}{\textit{Country}} & \multicolumn{1}{c}{\textit{Continent}} \\
   & \multicolumn{1}{c|}{} & \multicolumn{1}{c}{\textbf{1 km}}   & \multicolumn{1}{c}{\textbf{25 km}} & \multicolumn{1}{c}{\textbf{200 km}} & \multicolumn{1}{c}{\textbf{750 km}}  & \multicolumn{1}{c}{\textbf{2,500 km}}  \\ \hline
\multirow{10}{*}{\textbf{IM2GPS} \cite{hays2008im2gps}}   
   & PlaNet \cite{weyand2016planet}  & 8.4 & 24.5 & 37.6 & 53.6  & 71.3 \\
   & CPlaNet \cite{seo2018cplanet} & 16.5 & 37.1 & 46.4 & 62.0  & 78.5 \\
   & ISNs($M, f^*, S_3$) \cite{muller2018geolocation} & 16.9 & 43.0 & 51.9 & 66.7  & 80.2 \\
   & Translocator \cite{pramanick2022world}  & 19.9 & 48.1 & 64.6 & 75.6  & 86.7 \\
   & GeoDecoder \cite{clark2023we}    & 22.1 & \textbf{50.2} & \textbf{69.0} & 80.0  & 89.1 \\
   & PIGEOTTO \cite{Haas_2024_CVPR}      & 14.8 & 40.9 & 63.3 & \textbf{82.3}  & \textbf{91.1} \\
   & \textbf{Ours (InternVL2-76B)} & 22.1 & 49.7 & 62.8 & 76.3  & 89.8 \\
   & \textbf{Ours (Qwen2-VL-72B-Instruct)} & \textbf{23.2} & \textbf{50.2} & 62.8 & 78.0  & 90.7 \\ \cline{2-7} 
   & $\Delta$ (\% points) & \textbf{+1.1}  & 0 & -6.2  & -4.3 & -0.4   \\ \hline

\multirow{10}{*}{\textbf{IM2GPS3k} \cite{vo2017revisiting}} 
   & PlaNet \cite{weyand2016planet}  & 8.5 & 24.8 & 34.3 & 48.4  & 64.6 \\
   & CPlaNet \cite{seo2018cplanet} & 10.2 & 26.5 & 34.6 & 48.6  & 64.6 \\
   & ISNs($M, f^*, S_3$) \cite{muller2018geolocation} & 10.1 & 27.2 & 36.2 & 49.3  & 65.6 \\
   & Translocator \cite{pramanick2022world}  & 11.8 & 31.1 & 46.7 & 58.9  & 80.1 \\
   & GeoDecoder \cite{clark2023we}    & 12.8 & 33.5 & 45.9 & 61.0 & 76.1 \\
   & PIGEOTTO \cite{Haas_2024_CVPR}      & 11.3 & 36.7 & 53.8 & 72.4  & 85.3 \\
   & GeoCLIP \cite{vivanco2024geoclip}      & 14.1 & 34.5 & 50.7 & 69.7  & 83.8 \\
   & Img2Loc(GPT4V) \cite{zhou2024img2loc} & \textbf{17.1} & \textbf{45.1} & \textbf{57.9} & \textbf{72.9}  & 84.7 \\
   & \textbf{Ours (InternVL2-76B)} & 15.3 & 37.0 & 49.4 & 65.6  & 81.1 \\
   & \textbf{Ours (Qwen2-VL-72B-Instruct)} & \textbf{17.1} & 38.7 & 51.4 & 66.6  & \textbf{85.6} \\ \cline{2-7} 
   & $\Delta$ (\% points) & \textbf{0}  & -6.4 & -6.5  & -6.3 & \textbf{+0.9}  \\ \hline

\multirow{10}{*}{\textbf{YFCC4k}*  \cite{vo2017revisiting}}    
   & PlaNet \cite{weyand2016planet}  & 5.6 & 14.3 & 22.2 & 36.4  & 55.8 \\
   & CPlaNet \cite{seo2018cplanet} & 7.9 & 14.8 & 21.9 & 36.4  & 55.5 \\
   & ISNs($M, f^*, S_3$) \cite{muller2018geolocation} & 6.7 & 16.5 & 24.2 & 37.5  & 54.9 \\
   & Translocator \cite{pramanick2022world}  & 8.4 & 18.6 & 27.0 & 41.1  & 60.4 \\
   & GeoDecoder \cite{clark2023we}    & 10.3 & 24.4 & 33.9 & 50.0 & 68.7 \\
   & PIGEOTTO \cite{Haas_2024_CVPR}      & 10.4 & 23.7 & 40.6 & \textbf{62.2}  & \textbf{77.7} \\
   & GeoCLIP \cite{vivanco2024geoclip}      & 9.5 & 19.3 & 32.6 & 55.0  & 74.6 \\
   & Img2Loc(GPT4V) \cite{zhou2024img2loc} & 14.1 & 29.6 & 41.4 & 59.3  & 76.9 \\
   & \textbf{Ours (InternVL2-76B)} & 20.8 & 30.0 & 39.0 & 54.6  & 70.7 \\
   & \textbf{Ours (Qwen2-VL-72B-Instruct)} & \textbf{24.3} & \textbf{35.1} & \textbf{44.5} & 59.5  & 75.2 \\ \cline{2-7} 
   & $\Delta$ (\% points) & \textbf{+9.9}  & \textbf{+5.5} & \textbf{+3.1}  & -2.7 & -2.5   \\ \hline
\end{tabular}
\begin{tablenotes}
  \small
  \item * During the benchmarking process, we encountered an issue where 169 images out of the expected 4,536 were unavailable and returned the message ``This photo is no longer available.'' Thus, the benchmark was conducted using 4,367 images, which corresponds to 96.3\% of the dataset. This implies that 3.7\% of the images could not be analysed using our method.
\end{tablenotes}
\end{table*}

To assess the efficacy of our proposed method, we conducted comprehensive experiments on three benchmark datasets: IM2GPS \cite{hays2008im2gps}, IM2GPS3k \cite{vo2017revisiting}, and YFCC4k \cite{thomee2016yfcc100m}. We compared our approach against several methods, including PlaNet \cite{weyand2016planet}, CPlaNet \cite{seo2018cplanet}, ISNs \cite{muller2018geolocation}, TransLocator \cite{pramanick2022world}, GeoDecoder \cite{clark2023we}, PIGEOTTO \cite{Haas_2024_CVPR}, GeoCLIP \cite{vivanco2024geoclip}, and Img2Loc(GPT4V) \cite{zhou2024img2loc}. The evaluation metrics are based on the percentage of images localized within specific distance thresholds: 1 km (Street level), 25 km (City level), 200 km (Region level), 750 km (Country level), and 2,500 km (Continent level). Table \ref{tab:performance_comparison} summarizes the geolocation estimation accuracy across these methods and datasets.

In the Img2GPS benchmark dataset, our method, utilising the Qwen2-VL-72B-Instruct model, achieved a street-level (1 km) accuracy of 23.2\%, surpassing all prior approaches and establishing a new state-of-the-art at this level. At street-level accuracy, our method also ranked second, with the InternVL2-76B model and the GeoDecoder method achieving 22.1\%. Compared to the best-performing model in this category, our method represents an improvement of 1.1\%. At the city-level (25 km), our approach demonstrated exceptional performance, matching the GeoDecoder method to secure the best accuracy. On this benchmark dataset, our models achieved the highest overall accuracy of 50.2\%, correctly estimating the location for nearly one in two images. At the 200 km and 750 km distance error thresholds, our models exhibited competitive performance, although trailing the best results by 6.2\% and 4.3\%, respectively. Furthermore, at the continent level accuracy, our approach was among the two methods achieving over 90\% precision.

For the IM2GPS3k benchmark dataset, our approach achieved a street-level accuracy of 17.1\%, matching the state-of-the-art performance of Img2Loc (GPT4V), which also reached 17.1\%. While our method fell short of the highest accuracy at the city, region, and country levels by 6.4\%, 6.5\%, and 6.3\%, respectively, it excelled at the continent level, achieving an accuracy of 85.6\%. This was the highest among all compared methods, marking an overall improvement of 0.9\%.

In the YFCC4k benchmark dataset, our method demonstrated state-of-the-art performance at street, city, and region levels. Using the Qwen2-VL model, our approach surpassed the nearest competitor at the street level by 9.9\%, achieving an accuracy of 24.3\%, thereby setting a new state-of-the-art benchmark. Similarly, at the city level, our method achieved an accuracy of 35.1\%, outperforming the closest approach by 5.5\%. At the region level, it maintained state-of-the-art accuracy with a score of 44.5\%, exceeding the nearest competitor by 3.1\%. While the results at the country and continent levels were highly competitive, it is important to note that 169 images in the dataset remained inaccessible despite attempts from various sources. These images were excluded solely from our tests, accounting for 3.7\% of the dataset not being used in the analysis.

Across all benchmarks, our method demonstrates superior performance for geolocation estimation at the street level, which is the most ambitious level of granularity. The integration of MLLMs and RAG databases enables our method to leverage detailed visual and textual cues, enhancing its ability to make precise geolocation predictions. The delta values in Table \ref{tab:performance_comparison} represent the difference in percentage points between our method and the best methods. Positive delta values indicate an improvement over existing methods, and negative delta values at higher distance thresholds suggest areas for potential enhancement. Our method, In the IM2GPS benchmark dataset, there has been a +1.1\% improvement at the street level; in the IM2GPS3k benchmark dataset, a +0.1\% improvement at the continent level; and in the YFCC4k benchmark dataset, improvements of +9.9\% at the street level, +5.5\% at the city level, and +3.1\% at the region level.

\section{Discussion}

In artificial intelligence research, MLLMs, which excel in addressing generalization problems through training on extensive datasets, appear less proficient in geographic location estimation \cite{ligeoreasoner, wu2023mixed} without the application of RAG. However, as evidenced by our results, the integration of RAG significantly enhances their performance. Despite utilizing quantized weights rather than the originally trained weights of open-weight models, our approach achieved state-of-the-art accuracy at the street level — a critical metric for applications requiring precise geolocation — highlighting the method's effectiveness.

Our findings demonstrate that this refined approach achieves results that outperform existing methods across all metrics, including street-level and city-level granularity, without relying on costly pretraining or fine-tuning processes. This has been delivered by integrating a larger number of street-level photographs into the RAG database, employing a different image encoder, while prioritising open-source models. The consistency of our method's performance across diverse datasets and evaluation metrics underscores its robustness and generalizability.

It is worth noting, however, that the benchmark datasets used to compare the performance by this study, are subject to temporal inconsistency too. Therefore, some images remained inaccessible in a few cases and it is important to consider the potential impacts. Nevertheless,  this limitation did not significantly affect overall results, and our method demonstrated notable performance and resilience against these challenges, reaffirming its robustness when applied to such benchmark datasets.

\section{Conclusion}

This paper harnessed the power of MLLMs and the capabilities of RAG databases to achieve significantly improved performance in geolocation estimation accuracy at street level and other scales across multiple benchmark datasets. The superior results of our approach, compared to its predecessors following similar methodologies, can be attributed to three key factors: the inclusion of a substantially larger number of street-level images in the RAG database, the selection of SigLIP as the image encoder due to its superior performance metrics, and the preference for open-weight models, chosen for their reproducibility and open-source nature, despite not being the largest models available.

This demonstrates that solving a complex problem like geolocation estimation can be achieved with high precision, without the need for extensive model fine-tuning and retraining, thereby saving considerable time and resources. 

Our findings contribute valuable insights to the field, paving the way for the development of more efficient and accurate geolocation estimation tasks.

Future research, conducted with a focus on openly sharing resources such as datasets, open-weight models, and code, will undoubtedly enhance the reproducibility of these approaches. Furthermore, while fine-tuning such large models is perceived to be highly resource-intensive and costly, attempting it could yield intriguing results, given the strong language and vision capabilities of MLLMs.

\bibliographystyle{IEEEtranN} 
\bibliography{References}

\end{document}